# Creativity and Delusions:
# A Neurocomputational Approach

Daniele Q. Mendes
*LNCC*
Luís A. V. de Carvalho
*COPPE/UFRJ*

daniele@lncc.br    alfredo@cos.ufrj.br

*Abstract*

*Thinking is one of the most interesting mental processes. Its complexity is sometimes simplified and its different manifestations are classified into normal and abnormal, like the delusional and disorganized thought or the creative one. The boundaries between these facets of thinking are fuzzy causing difficulties in medical, academic, and philosophical discussions. Considering the dopaminergic signal-to-noise neuronal modulation in the central nervous system, and the existence of semantic maps in human brain, a self-organizing neural network model was developed to unify the different thought processes into a single neurocomputational substrate. Simulations were performed varying the dopaminergic modulation and observing the different patterns that emerged at the semantic map. Assuming that the thought process is the total pattern elicited at the output layer of the neural network, the model shows how the normal and abnormal thinking are generated and that there are no borders between their different manifestations. Actually, a continuum of different qualitative reasoning, ranging from delusion to disorganization of thought, and passing through the normal and the creative thinking, seems to be more plausible. The model is far from explaining the complexities of human thinking but, at least, it seems to be a good metaphorical and unifying view of the many facets of this phenomenon usually studied in separated settings.*

## I. Creativity

One of the most interesting and fuzzy of our mental activities is what we call creativity. Since Classical Antiquity, the act of creating new ideas, original artistic expressions, and unforeseen machinery has fascinated the philosopher and the layman. The mystery of creation seems to come from the fact that the "new" emerges from the "nowhere" of old, well-known, and current concepts.

Many have tried to define and partially explain the creative phenomenon. It could be, for Gagné, the combination of ideas from different and largely separated knowledge fields [1], or, second Rogers, the ability of making unusual relationships or unexpected connection between elements [2]. Associationists say that creative people are capable of linking external stimuli to highly unlikely answers, generating solutions masked for the majority [3]. Cognitivists explain creativity as another way of information processing or cognitive style [3].

Focusing attention on the central elements of a problem and disregarding the peripheral ones is a good strategy for finding a conventional and unique solution to a problem. This convergent-thought approach is naturally taught at schools and societies and used for the majority of the people in everyday life situations. However, broadening the attention to a wider range of elements and regarding them as potentially relevant may be a better approach to find out new and creative solutions. This divergent thought style follows many directions at the same time and allows the discovery of unusual associations of ideas.

For the Gestalt School, creativity is the reorganization of mental structures, producing new associations of ideas depending on the perception of the real situation [4]. The more flexible the mental reorganization, the more creative the thinking process.

Also, psychodynamical theories were proposed to explain creativity. Freud suggested that the creative act is a consequence of a fantastic view of the world when the real world frustrates someone's desires. If the incursion to fantasy does not alleviate the frustration, neurosis arises [3]. In the model of Adaptative Regression [5], the creative process is viewed as a regression to unconscious levels which allows a momentarily freedom from stereotyped and conventional scenes. Psychosis is seen as a involuntary and uncontrolled regression to childlike modes of thinking, while the creative person is capable of a temporary and controlled regressive trip.

Although inconclusive [6, 7, 8, 2, 9], psychodynamical theories gather in a single model creativity, psychopathology and unconsciousness. Indeed, many reports express a strong correlation between creative and psychotic thinking. In the seventies, creative writers and maniacs were compared and a common tendency to broaden or shift conceptual boundaries (overinclusion) was observed [10]. The overinclusiveness of the maniacs was based on bizarre associations while that of the writers was due to the recognition of original and valuable associations. In another study, schizophrenics and creative adults were tested and a common wider attentional focus was noted along with a capacity of making looser associations [11].

In the eighties, creativity and schizophrenic thought were suggested to be the same cognitive process based on the Alternate Uses Tests [12]. Recently, almost three hundred famous biographies were rated by the DSM III and creativity was again linked to pathological personality characteristics or disorders, mainly bipolar disorder [13]. Another study concluded that coarse rather than focused semantic activation is strongly related to schizophrenic thought and creative thinking [14]. Whatever the relation between psychopatology and creativity is, some commonalties seem to exist, like the idea of broader, distant or looser association making and unfocusing of attention. In the present paper, these commonalties will be explored to define an unifying model for creative and disturbed thought.

**II. Delusions**
Delusions are thought processes that deviate from the normal logical thinking by their character of subjective certainty, incorrigibility, and impossibility of content, as originally pointed out Karl Jasper [15]. Delusion is a primary phenomenon that express itself through judgments, and so, it is not the judgment indeed. That is why delusions cannot be understood and corrected even in the presence of many logical arguments. Indeed, as some delusions are possibly true, the impossibility of the content of a delusion was later changed to falsity. However, in some cases, like religious questions, true or falsity are not applicable [16]. Impossible, improbable, or even true, a delusion is a statement made in an inappropriate context or without a logical justification. Normal thought has reasons to justify itself and can embed the possibilities of doubt. Delusions are not followed by adequate and reasonable justifications and their property of total and unquestionable certainty leads to their incorrigibility.

Delusional manifestations are of three types: Delusional perception, representation and cognition. In the process of delusional perception, the patient attaches an abnormal meaning to a sensation or perception of the world. Usually, the real world perception is taken as a signal or revelation to the patient. In the delusional representation, a memory trace returns to consciousness with a new meaning while in the delusional cognition there are no perception or memory traces to attach new interpretations, but just an intuition that suddenly appears.

For Freud, delusion is a defence process where judgment mistakes are made when the ego tries to isolate from consciousness intolerable representations. When an intolerable idea is inseparably connected to reality, the only way of isolating it from consciousness is detaching from reality [17]. For the Gestalt School, some neurophysiological process breaks the coherence between perception and thinking, leading to the emergence of new "gestalts" [17]. Following the ideas of Hebb about cell assemblies [18], Fish [19] developed a neurobiological theory where the overstimulation of the cell assemblies that represent ideas of a sequential thought would lead to the process of delusion. In his theory, the reticular formation was the central responsible for the referred overstimulation., and the neurotransmitter serotonine was the neurochemical basis for the delusional thinking. Another important theory that relates delusions to the neurotransmitter dompamine will be reviewed in the next section.

Delusions begin from a mixture of anxiety, hiperarousal, suspicion, and the attachment of meaning to insignificant events. Once a meaning is attached, the patient will not question the event anymore and will further elaborate on it. This delusional work is an attempt to find coherence in his unusual thoughts. Acute delusions respond to neuroleptic treatment while chronic delusions tend to be resistant. Chronic delusions are not a state where the person is but part of the individual values, intentions, and views. It seems that the chronically deluded patient has a structural deformation that may have developed from the dynamical forces present in the acute delusion [20]. Chronic delusions may also develop from a state of sensorial deprivation like, for example, isolated individuals (prisoners, refugees, hearing loss).

**III. Dopaminergic Modulation**
The catecholamines norepinephrine, epinephrine, and dopamine are important neuroactive substances produced in some brain sites and released at distant and widespread areas in a diffuse or divergent way [21]. These substances do not act through membrane ion channels but, instead, activate intracellular messengers, promoting a longer effect than the other neurochemicals released by synapses inside the brain. As these other chemicals have specific and local synaptic patterns, act through ion channels, and have short-lasting effects, it is interesting to suppose that they differ from the cathecolamines in function. Indeed, the substances released by synapses in the CNS may be classified in the two broad categories of neurotransmitters and neuromodulators [22]. Due to the fast action and connection patterns of their producing synapses, neurotransmitters seem to be involved in the immediate processing of signals, while the neuromodulators, with their opposing properties, hint for a regulatory function, modulating the operational characteristics of the receptor neurons, i.e. their responses to neurotransmitters [23].

Increases or decreases in the catecholaminergic levels have behavioral consequences in arousal, attention, learning, memory, and motor responses [24]. It is not clear, but it seems plausible to suppose that catecholamines affect the neuronal ability to discern what is information from what is noise in a signal. Some authors suggest that these neuromodulators enhance the stronger signal and dampen the weak one [20], while others advocate that they enhance the cell sensitivity to either excitatory and inhibitory signals [22]. Whatever the mechanism is, the net effect is the enhancement of the signal in relation to the background, spontaneous activity called "noise." The signal-to-noise ratio at neuronal level has been associated with the performance in some cognitive tasks and behaviors, including the deviant behavior of psychosis.

The dopamine hypothesis of schizophrenia advocates that the disorder is caused by an overactivity of the brain dopaminergic system [20]. Observations that dopamine antagonists alleviate schizophrenic acute symptoms support the hypothesis [25]. An elaboration of this hypothesis is that the dopamine release is chronically reduced in schizophrenic patients, leading to the upregulation of the postsynaptic receptors and a consequent intensified response in moments of normal or increased dopamine release, for example, due to environmental stressors [26]. This would explain both, the positive and the negative, symptoms of the disease.

A relation between acute delusions and dopamine activity is clear from the fact that amphetamine can cause psychotic states with paranóia, hyperarousal, hyperactivity, and suspiciousness. It also seems that a decreased dopamine level leads to a higher signal-to-noise ratio and looser associations of thought, allowing the creation of new relations [20]. For example, overinclusion and semantic priming are two phenomena observed in schizophrenic patients that can be related to lower dopamine levels and to abnormally looser thought associations [24].

**IV. Cortical Maps**
In the middle of the 19th century the scientists Helmholtz and Mach studied many phenomena of the visual perception in humans. Particularly, they were interested in optical illusions like the fact that edges or contours between light and dark parts of an image tended to be enhanced in relation to the light and dark interior of the image. They explained the illusion hypothesizing that in the human retina the cells are excited by light that converges to a central region and are inhibited by the light that projects to the surrounding areas. Almost a century later, experimental results showed that the eye of the crab called *Limulus* [27] and some vertebrates [28] have an structure, then called on-center/off-surround, in which a neuron is in cooperation, through excitatory synapses, with the neurons in the immediate neighborhood while it is in competition with the neurons which lay outside these surroundings. There is experimental evidence supporting that the same mechanism is also present in the mammalian central and peripheral nervous system. It seems that pyramidal cortical cells are connected in this on-center/off-surround way [29]. Other areas in the brain, like the hippocampus and the cerebellum show the referred hardwired structure [30, 31].

Competition and cooperation are found not only statically hardwired but also as part of many neuronal dynamical processes. As a matter of fact, competition is essential to the neurodevelopment where neurons compete for certain chemicals. In synaptogenesis, for example, the substances generically called *neural growth factors* are released by stimulated neurons and, spreading through diffusion, reach the neighboring cells, promoting synaptic growth. Cells that receive neural growth factors make synapses and live, while the cells that have no contact with these substances die [32]. A neuron that releases neural growth factor guides the process of synaptic formation in its tridimensional neighborhood, becoming a center of synaptic convergence. When some neighboring neurons release different neural growth factors in different amounts, many synaptic convergence centers are generated and a competition is established between them by the synapses of their surroundings. It seems that at least two processes participate in the dynamics of synaptic formation: pre-synaptic neurons competing for neural growth factors to survive and pos-synaptic neurons that release neural growth factors competing for synapses that will keep them alive with stimuli. It is worth noting that, as a single neuron is capable of receiving and releasing neural growth factors at the same time, the two competition processes described above effectively occur in every neuron and, consequently, a signaling network is established to control the development and plasticity of neural circuits. Remembering that all this competition is started and controlled by environmental stimulation, it is possible to have a glimpse to the way the environment records or represents itself in the brain.

The competition processes described above are essential to the formation of some neuronal organizations called *maps*. A neural map is a biological circuit composed of two sets of neurons, called domain and image, in such a way that similar patterns of activation of the domain are projected to neighboring neurons in the image. In other words, a neural map is a projection that transfers similarities at the domain to spatial relationships at the image. Maps were first observed in 1937 [33] and later the concept was refined [34] taking the somatosensory and motor cortices as models. Studies of the visual [35], somatosensory [36], and associative [37] cortices showed that small regions of those tissues respond to similar stimuli. Indeed, stimuli like position, orientation, color, spatial frequency, auditory frequency, and also meanings [38, 39, 40, 20] are capable of being represented in neuronal circuits as maps.

These maps are subject to constant change, not only in the neurodevelopmental phase, but throughout life as a function of one's experiences [41]. Monkeys trained to discriminate between two different vibrations imposed to the finger skin showed an increase in the region of the somotosensory cortex responsible for the finger representation [42]. Marked cortical changes were also demonstrated in blind subjects when comparing the braille reading finger cortical representation to the other fingers representations [43].

Maps have puzzled neuroscientists in the last decades, mainly the question of how do they arise from the simple on-center/off-surround wiring pattern. Computational theories gave some important insights to the problem, since some cortical maps are artificially developed from simple governing rules of synaptic plasticity in computer simulation models. The most general of these models is called the *Self-Organizing Map* [44] in which two sheets of neuronal tissue with *n* neurons each, corresponding to the domain and the image, are initially randomly connected in a way that every neuron *i* at the image receives synaptic projections $w_i \in R^n$ from every neuron at the domain (Figure I). Neurons at the domain don't form synapses among themselves and receive "sensory" inputs (stimuli), while neurons at the image make synapses following the on-center/off-surrounding paradigm, i.e., short-range excitation or cooperation and long-range inhibition or competition.

The on-center/off-surround synapses don't change during the development of the map, while the synapses between the domain and the image are modified along the process of map formation. Indeed, every time the neural network is in contact with a stimulus $x_k \in R^n$, $k=1,2,...$ in its domain, there will be only one excited neuron *i\** at the image because the short-range cooperation and long-range competition makes the more excited neuron inhibit the others. The position *r\** of this winner neuron at the image determines how much the synapses will be modified. Synapses from neurons closer to the winner will be strongly changed in such a way that these neurons will be more intensely excited by the stimulus $x_k$ in a next time. Synapses from neurons distant from the winner will be weakly changed or not changed at all, depending on the dispersion $\sigma$ of the neighborhood function $\phi(r_i, r^*)$, where $r_i \in R^n$ gives the position of a neuron $i$ at the image sheet. By this process, every neuron in the image will be more easily excited by the stimulus $x_k$ (synaptic facilitation) in the future. The development of the map is due to the fact that the amount of synaptic facilitation is proportional to the distance from the winner neuron. The process of synaptic modification $\Delta w_i^l$ for each neuron $i$ is repeated for every learning step $l$ where the stimulus $x_k \in R^n$, $k=1,2,...$ is presented to the neural network, and is given by

$$\Delta w_i^l = \rho(l) \cdot \phi(r_i, r^*) \cdot (x_k - w_i), \qquad (1)$$

where $\rho(l)$ is the learning rate defined by

$$\rho(l) = \rho_0 \cdot \beta^{(l-1)}; \; 0 < \beta < 1, \; l=1,2,... \qquad (2)$$

The learning rate begins with the value $\rho_0$ and decreases with the learning step $l$ with a rate $\beta$.

The neighborhood symmetric function $\phi(r_i, r^*)$ takes the form of a gaussian function like

$$\phi(r_i, r^*) = exp -( \| r_i - r^* \|^2 / 2 \, \sigma(l)^2 ). \qquad (3)$$

The initial dispersion of the gaussian, $\sigma_0$, is high, representing that all the neurons in the image are considered neighbors. This allows the modification of the randomness of the initial synapses to a more organized pattern where neighborhood is of capital importance. Every time step $l$ that another stimulus is presented to the neural network domain, the neighborhood shrinks a bit, gradually giving to the map a local organization. The dispersion $\sigma(l)$ at each learning step is given by

$$\sigma(l) = \sigma_0 \cdot \alpha^{(l-1)}; \; 0 < \alpha < 1, l=1,2,... \qquad (4)$$

where $\alpha$ is a decrement rate.

The way the learning rate decreases and the neighborhood shrinks is fundamental to the map development. A faster decrement in the learning rate does not give enough time to the synapses to change, and so the randomness of the initial synaptic pattern is consolidated at the end of the process. When neighborhoods shrink rapidly, the level of neuronal cooperation necessary to produce maps are not present and neighborhood relationships are ill-defined at the end of the simulation. Indeed, the neighborhood function may be likened to the steady-state concentration profile of a neural growth factor in the neural tissue. When the dynamical equilibrium between neural growth factor release and metabolization is accomplished in every region of the tissue, due to the diffusion process, a concentration profile that asymptotically decreases with radial distance is attained (see Figure III). The parameter $\sigma_0$ represents the amount of neural growth factor released by the neurons at the beginning of the neurodevelopment process.

As plasticity is always happening in our brains, if the parameter $\sigma$, that controls the rate of synaptic alteration, is kept constant, the map will represent a cortex which is capable of changes during one's entire lifetime.

**V. Simulation Results**

A self-organizing neural network with its two bi-dimensional sheets composed of 400 neurons each was developed for computer simulation, as shown in Figure I. A set of different stimuli, symbolized by the geometrical markers and representing different concepts or ideas, was repeatedly presented to the Domain sheet of the neural network. Due to the existence of feedforward connections between the Domain and the Image sheet, every stimulus presented to the Domain is projected to the Image. Initially, as the synapses are randomly generated, the stimuli presented to the Domain sheet are projected to random positions at the Image layer. As long as the stimuli are repeatedly presented to the neural network, the synapses change and a map-like structure develops at the Image layer. Similar stimuli, representing nearly associated or similar concepts, when presented to the Domain layer, lead to the excitation of neighboring regions in the Image neuronal layer. The contrary also holds as different stimuli, representing dissimilar or not directly associated concepts or ideas, when presented to the self-organizing neural network will excite neurons at distant regions at the Image sheet. This is what we call a semantic map.

The purpose of our simulations is to show that different maps arise when dopaminergic modulation controls the synaptic formation process. In fact, varying the parameters responsible for the signal-to-noise ratio results in maps that represent the concepts or ideas in a way that can be likened to the delusional, creative, and disorganized thought. To simulate the signal enhancement promoted by the dopaminergic activity, a threshold $\theta$ is associated to every neuron at the Image sheet [22]. When the total signal input, coming from the Domain layer to an Image sheet neuron, exceeds the threshold, this neuron is considered to be excited. Increasing or decreasing the threshold will promote the effect of dopaminergic enhancement or dampening of the incoming signal. The simulation of noise is simply obtained by adding to the stimulus a random number with a range between $-p$ and $+p$ where $p$ is a percentage of the stimulus value [22]. The parameters $\theta$ and $p$ allow us to realize any simulation desirable with total control flexibility over the signal-to-noise ratio.

In a first simulation experiment, a semantic map was allowed to develop from the self-organizing neural network when ten stimuli, representing ten different concepts or ideas, were repeatedly presented to the Domain layer with no noise and a predefined signal level $\theta$ of 0.999. This map is represented in Figure I and will stand as a reference for future comparisons. Note that the similar stimuli * and ✳ were mapped into neighboring regions of the Image neuronal layer while, for example, the very different stimuli represented by / and + were mapped into the opposing corners of the Image sheet. This observation was done just to show that the map was well-formed for these ten concepts or ideas. Now, in a second simulation, the Domain sheet of this already well-formed map will be excited by the single stimuli represented with an *. The dopaminergic modulation was changed in this simulation with the addition of a noise level $p$ of 10 %. The resulting Image layer map can be seen in Figure II. Note that with the addition of noise, the stimulus * expanded its representation, exciting neurons outside its original region at the Image layer and invading the region represented by the concept represented by ✽. This can be interpreted as if the increase of noise level, or equivalently the decrement of the signal-to-noise ratio, was capable of promoting the association of the different, but similar, ideas or concepts * and ✽, neighbors in the map. Much of our reasoning can be understood as an association of ideas. Indeed, when a stimulus (endogenous or exogenous) elicits a central idea, that we will call here a "thesis", other ideas, that corroborate or refute the thesis, are spontaneously elicited. Let us call these spontaneously elicited ideas of "antitheses." As the thesis and the antitheses are elicited at the same time, they are temporally associated, and the final result of this simultaneous presence is the weighted sum of their influences, emerging a final pattern that we will call here the "synthesis" or the conclusion of the reasoning process. If we assume that the "normal" thought is the triggering of a thesis that elicits a group of antitheses which will be weighted (pondered) together to generate a synthesis, then, for the occurrence of the "normal" thought, it is necessary some level of noise or a relatively lower dopaminergic modulation of the signal-to-noise ratio.

In the next simulation, the noise level will be increased from 10 % to 170 % and the same procedure realized in the second experiment repeated. The result is shown in Figure III. Note that now the central stimulus * (thesis) has excited many neurons outside its original representation, invading areas where others stimuli were represented. In our model, this means that a central idea (thesis) has been associated with many other ideas (antitheses) generating a pattern that we can liken to the creative thinking. If, in the "normal" thought, a central idea (thesis) is associated to a few neighboring and similar ideas (antithesis), in the creative thinking, this same central idea, will be associated to different, normally not associated, ideas. The process of making associations between a central stimulus and distant ones resembles the formerly reviewed theories of creativity where concepts like "loosening of associations", "divergent-thought", "the ability of making unusual relationships", "flexibility of mental organization", "a momentarily freedom from stereotyped and conventional scenes", "the broadening of the conceptual boundaries", "the unfocusing of attention", and some other similar concepts are always present. As a consequence, to the occurrence of creative thinking, it is necessary a higher level of noise, or equivalently, a lower dopaminergic modulation of the signal-to-noise ratio, as experimentally observed [24].

The same way that coarse rather than focused semantic activation is strongly related to schizophrenic thought and creative thinking [14], the model presented here can show the subtle border between creativity and disorganized thought. Indeed, if the signal-to-noise dopaminergic modulation is still more reduced as the consequence of an increase in the noise level $p$ from 170 % to 200 %, and the same simulation experiment repeated, a new pattern will appear in the Image layer of the neural network, as can be seen in Figure IV. Note that this increase in noise was sufficient to make the same stimulus * invade other distant areas that it had not invaded in the anterior simulations. This means that the central idea (thesis) elicits a plethora of other ideas (antitheses) resulting in a new pattern that represents a synthesis where all the ideas are present and associated between themselves. It is not possible in this case to know what is the central idea and what is laterally associated. The synthesis lacks a coherence in relation to the thesis because all the associated ideas are equally present and weighted, and opposing and corroborating ideas have the same influence over the conclusion (synthesis). The synthesis encompasses any idea independently of its contents or proximity in relation to the thesis. We can say that a synthesis like this represents a disorganized thought that follows no direction or have no consistent meaning. In other words, when noise is higher, the association of ideas becomes more flexible and the creative thought degenerates to disorganization. The border between creativity and disorganization is obviously not clear as seen in the results reviewed at the beginning of this paper. As a consequence, the level of dopaminergic modulation of the signal-to-noise ratio that split the geniality from the illness can not be determined. Actually, the model has shown up to now that there is a continuum ranging from the normal thought to the disorganized one, passing through what we call creativity. In the next simulation, this continuum will be shown to encompass even the delusional thought.

As dopamine regulates the signal-to-noise ratio, it is necessary for realistic simulations to calibrate these two important variables, signal and noise, to generate values for this ratio that are significant to our experiments. In the previous simulations, the noise was gradually increased promoting the association between a central idea and more and more distant concepts. In the next and last experiment, the noise will be kept constant at a value $p$ of 5 % and the signal level will be increased from 0.999 to 0.9995. The same ten stimuli were presented to the neural network and the central idea , *, had its signal level increased. The Domain layer and the neurons excited at the Image sheet are shown in Figure V. Note that, in comparison with the original map described in Figure I, the area occupied by the ideas has shrunk. This shrinking process make the representation more focused and the associations between the ideas represented more unlike to occur. The stimulation of the neural network with an stimulus representing an idea (thesis) probably will not elicit the concomitant excitation of neighboring ideas (antitheses) because the shrinking process has separated the regions from one another. In this situation, the synthesis becomes equal to the thesis because there are no antitheses to corroborate or refute the central idea (thesis). The "normal" thought process of weighting many ideas with different influences to achieve a conclusion does not happen any more. It is possible to liken this phenomenon with the delusional thinking because the absence of antitheses does not allow the embedding of doubts, resulting in the character of unquestionable certainty and incorrigibility of delusions. This last simulation shows that the model unifies the many-faced phenomenon of normal and abnormal thinking. Different thinking processes are viewed just as possible positions over a one-dimensional continuum where the signal-to-noise ratio is the measure. At one extreme of this line, where the signal-to-noise ratio is high, the semantic map becomes more focused in the representations of ideas, resulting in the delusional thinking. At the other end of the linear continuum, where the signal-to-noise ratio is low, the excessive noise promotes unusually associations between ideas resembling the disorganized thought. The "normal" and the creative thought processes are positioned between these two ends, depending on the noise level, as can be pictorially shown in Figure VI.

## VI. Conclusions

Based on experiments that hint to the dopaminergic signal-to-noise modulation of the CNS neurons, and hypothesizing the existence of semantic cortical maps that would represent concepts or ideas, a self-organizing neural network model was developed to unify the different thought processes in a single neurocomputational substrate. Simulations were performed varying the two principal control parameters of the dopaminergic modulation which are the signal and the noise levels carried by the neurons from the input to the output of the neural network. Stimuli representing different ideas or concepts were mapped in a self-organized way and this map was taken as a reference for the other simulations. These simulations were performed simply by stimulating the neural network input layer with a single stimulus and observing the areas of the output layer excited. At each simulation, the signal-to-noise ratio was varied and different patterns emerged at the output layer. Basically, the stimulus used in the stimulation of the input layer was compared to a trigger of a central idea (or a thesis) at the output layer that, depending on the signal-to-noise ratio, invaded or not the neighboring areas that represented other ideas (antitheses). Assuming that the thought process (or synthesis) is the total pattern elicited at the output layer of the neural network as the result of the weighted influence of every area (thesis and antitheses) excited, the model could show how the "normal" and "abnormal" thinking are generated. In addition, it was shown that the borders between the different thought processes ("normal" or "abnormal") are fuzzy because, actually, there are no borders, but a continuum. The transition from a high signal-to-noise ratio to a low one results in a qualitative change of the reasoning process, ranging from delusion to disorganization of thought, and passing through what we may call the "normal" and the creative thinking. The model unifies the qualitative different thinking processes into a neurobiologically-based substrate and also shows that these processes define a continuum with gray zones where their differentiation is difficult or impossible. Although biologically plausible and experimentally based, the model is far from explaining the complexities of human thinking but, at least, it seems to be a good metaphorical and unifying view of the many facets of this phenomenon usually studied in separated settings.

**FIGURES**

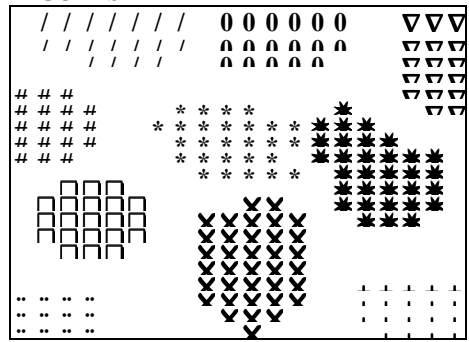

Figure I - A reference map with ten different concepts represented on it.

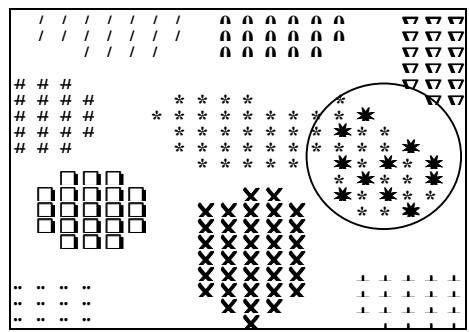

Figure II - The central idea * (thesis) is associated with a neighboring idea ✷ (antithesis), leading to the formation of a pattern that is the conclusion of the thinking process, or the synthesis.

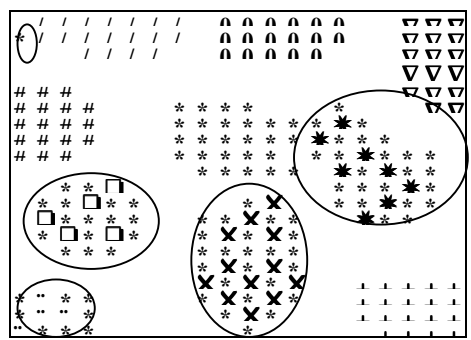

Figure III - The central idea * (thesis) is associated with distant ideas (antithesis), leading to the formation of a pattern that can be likened to the creative thinking.

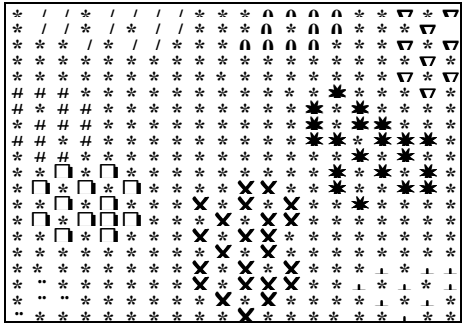

Figure IV - The central idea * (thesis) is associated with all ideas, leading to the formation of a pattern that can be likened to the disorganized thought.

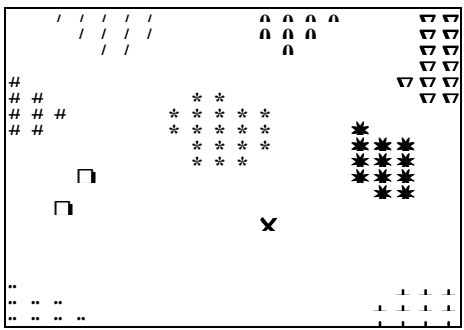

Figure V - The stimulation of the neural network with a higher signal level leads the ideas to shrink their original region in the Image layer, hindering their association. Without associations, the synthesis becomes the thesis and the antitheses are not considered or pondered. This map seems to represent a rigidity of thought or a delusional thinking.

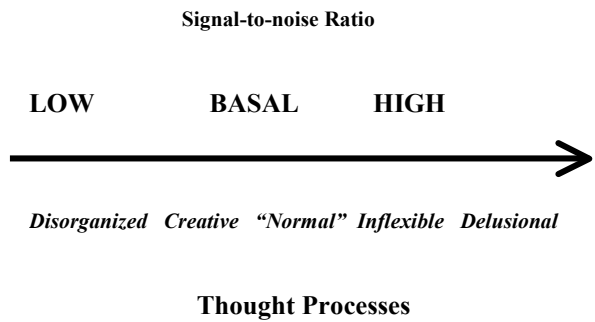

Figure VI – The linear unifying continuum of thought processes based on the signal-to-noise dopaminergic modulation.

**References**
1. Novaes, M. 1971, *Psicologia da Criatividade*. Petrópolis, Ed. Vozes.
2. Rogers, C., 1949, "Towards a Theory of Creativity". In: Anderson, H. (ed), *Creativity and its Cultivation*. New York, Harper & Brothers Publishers.
3. Franchi, L. *Delimitação do Conceito Atual de Criatividade*. 1972, Dissertação de Mestrado, Pontifícia Universidade Católica, Rio de Janeiro, RJ, Brasil.
4. Wertheimer, M., 1968, *Productive Thinking*. London, Tavistock.
5. Kris, E., 1952, *Psychoanalytic Explorations in Art*. New York, International Universities Press.
6. Schachtel, E., 1959, *Metamorphosis: On the Development of affect, perception, attention and Memory*. New York, Basic Books.
7. Maslow, A., 1962, *Toward a Psychology of Being*. Princeton, Van Nostrand.
8. Coleman, J., 1960, *Personality: Dynamics and Effective Behavior*. Chicago, Scott, Foreman & Co.
9. Hebeisen, A., 1960, "The Performance of a Group of Schizophrenic Patients on a Test of Creative Thinking". In: Torrance, E. (ed), *Creativity: Second Minnesota Conference on gifted Children*, Minneapolis, Minnesota Centre for Communication Study.
10. Andreasen, N., Powers, P., 1976, "Creativity and Psychosis", *Archives of General Psychiatry,* v. 32, pp. 70-73.
11. Dykes, M., 1976, "A Comparative Study of Attentional Strategies of Schizophrenic and Highly Creative Normal Subjects". *British Journal of Psychiatry*, v. 128, pp. 50-56.
12. Keefe, J., Magaro, P., 1980, "Creativity and Schizophrenia: An Equivalence of Cognitive Processing". *Journal of Abnormal Psychology*, v. 89, n. 3, pp. 390-398.
13. Post, F., 1994, "Creativity and Psychopathology: A Study of 291 World-Famous Men". *British Journal of Psychiatry*, v. 165, pp. 22-34.
14. Leonhard, D., Brugger, P., 1998, "Creative, Paranormal, and Delusional Thought: a Consequence of Right Hemisphere Semantic Activation?". *Neuropsychiatry, Neuropsychology, and Behavioral Neurology*, v. 11, n. 4, pp. 177-183.
15. Jaspers, K., 1966, *Psicopatologia Geral*. Buenos Aires, Ed. Beta.
16. Janzarik, W., 1988, *Strukturdynamishe Grundlagen der Psychiatrie.* Stuttgart, Enke.
17. Muller, E. *A Metapsicologia de Freud como uma Neuropsicologia*. 1976, Dissertação de Mestrado, Pontifícia Universidade Católica, Rio de Janeiro, RJ, Brasil.
18. Hebb, D., 1949, *Organization of Behavior*. New York, Wiley.
19. Fish, F., 1973, "A Neurophysiological Theory of Schizophrenia". In: Mahler, B. (ed), *Contemporary Psychology*, England, Penguin Books.
20. Spitzer, M., 1995, "A Neurocomputational Approach to Delusions". *Comprehensive Psychology*, v. 36, n. 83, pp. 83-105.
21. Molinoff, P., Axelrod, J., 1971, "Biochechemistry of Catecholamines". *Ann. Ver. Biochemistry*, pp. 465-500.
22. Servan-Schreiber, D., Printz, H., Cohen, J., 1990, "A Network Model of Catecholamine Effects:

Gain, Signal-to-Noise Ratio, and Behavior". *Science*, v. 249, pp. 892-895.
23. More, R., Bloom, F., 1978, "Central Catecholamine Neuron Systems: Anatomy and Physiology of the Dopamine Systems". *Ann. Rev. Neuroscience*, v. 1, pp. 129-169.
24. Spitzer, M., 1997, "A Cognitive Neuroscience View of Schizophrenic Thought Disorder". *Schizophrenia Bulletin*, v.23, n. 1, pp. 29-50.
25. Davis, K., Kahn, R., Ko, G., et al., 1991, "Dopamine in Schizophrenia: A Review and Reconceptualization". *American Journal of Psychiatry*, v. 148, n. 11, pp. 1471-1486.
26. Grace, A., 1991, "Phasic Versus Tonic Dopamine Release and the Modulation of Dopamine System Responsivity: A Hipothesis for the Etiology of Schizophrenia". *Neuroscience*, v. 41, pp. 1-24.
27. Hartline, H., Ratliff, F., 1957, "Inhibitory Interactions of Receptor Units in the Eye of Limulus". *Journal of General Physiology*, v. 40, pp. 351-376.
28. Kuffler, S., 1953, "Discharge Pattern and Functional Organization of Mammalian Retina". *Journal of Neurophysiology*, v. 16, pp. 37-68.
29. Szentagothai, J., 1967, "The Module Concept in Cerebral Cortex Architecture". *Brain Research*, v. 95, pp. 475-496.
30. Andersen, P., Gross, G., Lomo, T., Sveen, O., 1969, *Participation of Inhibitory and Excitatory Interneurones in the Control of the Hippocampal Cortical Output*. S. Francisco, Univ. of California Press.
31. Eccles, J., Szentagothai, J., 1967, *The Cerebellum as a Neuronal Machine*. New York, Springer.
32. Kandel, E., 1991, "Cellular Mechanisms of Learning and the Biological Basis of Individuality". *Principles of Neuroscience*. Norwalk, Appleton and Lange.
33. Penfield, W., Boldrey, E., 1937, "Somatic Motor and Sensory Representation in the Cerebral Cortex of Man as Studied by Electical Stimulation". *Brain*, v. 60, pp. 389-343.
34. Penfield, W., Rasmussen, T., 1950, *The Cerebral Cortex of Man: A Clinical Study of Localization and Function*. New York, Macmillan Press.
35. Hubel, D., Wiesel, T., 1965, "Receptive Fields and Functional Architecture in Two Non-Striate Visual Areas (18 and 19) of the Cat". *Journal of Neurophysiology*, v. 28, pp. 229-298.
36. Mountcastle, V., 1957, "Modality and Topographic Properties of Single Neurons of Cat's Somatic Sensory Cortex". *Journal of Neurophysiology*, v. 20, pp. 408-434.
37. Goldman-Rakic, P., 1984, "Modular Organization of the Prefrontal Cortex and Regulation of Behavior by Representional Memory". *Trends in Neuroscience*, v. 7, pp. 419-429.
38. Robson, J., 1975, "Receptive Fields: Neural Representation of the Spatial and Intensive Attributes of the Visual Image". *Handbook of Perception*. v. 5, New York, Academic Press.
39. Reale, R., Imig, T., 1980, "Tonotopic Organization in Auditory Cortex of the Cat". *Journal of Comprehensive Neurology*, v. 192, pp. 265-291.
40. Ritter, H., Kohonen, T., 1989, "Self-Organizing Semantic Maps". *Biol. Cybern.*, v. 61, pp. 241-254.
41. Mezernich, M., Sameshima, K., 1993, "Cortical Plasticity and Memory". *Current Opinion in Neurobiology*, v. 3., pp. 187-106.
42. Recanzone, G., Jenkins, W., Hradeck, G., et al., 1992, "Progressive Improvement in Discriminative Abilities in Adult Owl Monkeys Performing a Tactile Frequency Discrimination Task". *Journal of Neurophysiology*, v. 67, pp. 1015-1030.
43. Pascual-Leone, A., Torres, F., 1993, "Plasticity of the Sensorimotor Cortex Representation of the Reading Finger in Braille Readers". *Brain*, v. 116, pp. 39-52.
44. Kohonen, T., 1982, "Self-Organized Formation of Topologically Correct Feature Maps". *Biological Cybernetics*, v. 43, pp. 59-69.